Tech Science Press

# Big Data Testing Techniques: Taxonomy, Challenges and Future Trends

**Iram Arshad[1,*], Saeed Hamood Alsamhi[1] and Wasif Afzal[2]**

[1]SRI, TUS, Athlone, Ireland

[2]Mälardalen University, Västerås, Sweden

[*]Corresponding Author: Iram Arshad. Email: i.arshad@research.ait.ie

**Abstract:** Big Data is reforming many industrial domains by providing decision support through analyzing large data volumes. Big Data testing aims to ensure that Big Data systems run smoothly and error-free while maintaining the performance and quality of data. However, because of the diversity and complexity of data, testing Big Data is challenging. Though numerous research efforts deal with Big Data testing, a comprehensive review to address testing techniques and challenges of Big Data is not available as yet. Therefore, we have systematically reviewed the Big Data testing techniques' evidence occurring in the period 2010-2021. This paper discusses testing data processing by highlighting the techniques used in every processing phase. Furthermore, we discuss the challenges and future directions. Our findings show that diverse functional, non-functional and combined (functional and non-functional) testing techniques have been used to solve specific problems related to Big Data. At the same time, most of the testing challenges have been faced during the MapReduce validation phase. In addition, the combinatorial testing technique is one of the most applied techniques in combination with other techniques (i.e., random testing, mutation testing, input space partitioning and equivalence testing) to find various functional faults through Big Data testing.

**Keywords:** Big Data; Testing techniques; Testing process

## 1 Introduction

Big Data refers to datasets whose size is beyond the ability of typical database software tools to capture, store, manage and analyze [1]. Big Data can analyze for insights that lead to more informed decisions and strategic business decisions. Advanced technologies play a vital role in capturing, storing, managing and analyzing to get the value from data, i.e., Hadoop, Spark, Hive, Non-rational Structure Query Language (NO-SQL), cloud, edge computing, etc. Furthermore, the data volume and the importance of information extracted from data is not limited to large government agencies, large enterprises, or Internet Websites. From government agencies to an ample variety of organizations, ranging from small to large enterprises, are dealing with a flood of data. The value of Big Data is determined by how the company, organization and smart factories use the information obtained. Utilizing Big Data provides better customer service, enhances operations, generates tailored marketing campaigns, improves decision-making accuracy and takes other activities. In smart manufacturing, Big Data aids in the integration of previously fragmented systems, allowing businesses to understand their production processes better while automating data gathering and analysis. Therefore, the challenges with data in manufacturing systems and business have changed from gathering a significant amount of data to figuring out how to make the most use of the increasingly huge volumes of data accessible to make better business choices.

Data growth sources that drive Big Data technology investment are vast. Some represent entirely new data sources, while others change the resolution of existing data generated. New data sources for Big Data include industries that just recently began to digitize their content. Data growth rates in the last couple of years have been nearly infinite since they started from zero in most cases, i.e., healthcare, media/entertainment, life sciences, video surveillance, transportation, logistics, retail, utilities and telecommunications. Millions of IoT devices such as sensors and different smart devices are deployed on Internet of Things (IoT) networks for various applications. Different techniques and methodologies have been proposed to collect the data from these sources. For example, light-weight Unmanned Aerial Vehicles (UAVs) devices are used to collect data from different IoT devices [2]. Furthermore, social media such as Facebook, Twitter and Foursquare are the new data sources [3]. For example, recent stats show that in the first quarter of 2022, Facebook has roughly 2.93 billion active users per month [4]. The merging of the actual and virtual worlds is industry 4.0. This digital revolution is characterized by technology that harnesses Big Data and Artificial Intelligence (AI) to develop self-learning systems. In order to achieve manufacturing excellence, today's manufacturers seek business information through the compilation, analysis and exchange of data across all critical functional areas.

Many different terminologies are used to represent massive datasets. The authors in this study [5] introduced a difference between Big Data, Big Data Analytics (BDA) and Data Analytics (DA). For instance, BDA plays a vital role in Industry 4.0, which includes sensor data from manufacturing equipment that are examined to anticipate when maintenance and repair activities are required. As a result, manufacturers may improve production efficiency, better understand their real-time data and automate production management using self-service platforms, predictive maintenance optimization and automation. BDA in smart manufacturing is in predictive maintenance operations and sends out repair warnings and preventative maintenance, preventing equipment failures before they happen. In addition, sensors for condition-based monitoring can also be incorporated to monitor equipment performance and health in real-time, increasing total factory floor equipment efficiency. As a result, manufacturers perceive fewer failures, higher equipment productivity and more dependability due to decreasing failures and enabling better preventative maintenance procedures, all of which enhance total equipment effectiveness measures. In addition, BDA is an interdisciplinary field that includes many other scientific disciplines, such as computational intelligence, statistics, machine learning, signal theory, pattern recognition, operations research, predictive analytics, data mining, artificial intelligence, natural language processing, business intelligence, prescriptive analytics and descriptive analytics. Similarly, DA deals with visualization, cloud computing, or data exploration [5].

Big Data could be structured, semi-structured and unstructured, collected from different sources used for further processing to turn the data into a meaningful and valuable form. Much publicized and intact data attributes are known as V's of Big Data. These are 3V's volume, velocity and veracity [6] complemented with variability and value [7]. Later, veracity and visualization are included as additional quality characteristics of Big Data [8]. In Fig. 1, visual representation of Big Data is presented. From the diagram, we can observe the importance of the Big Data pipeline. BDA, DA and business intelligence pipelines rely on Big Data pipelines. If Big Data testing does not perform well, it has a significant adverse effect on all other pipelines; thus, it puts a lot of pressure on the test team to effectively prepare for Big Data testing. However, testing Big Data is challenging due to verifying and validating these data attributes. Therefore, Big Data testing is characterized as an examination of the Big Data pipeline. The literature has observed that Big Data testing is one of the biggest challenges faced by organizations dealing with Big Data. The essential challenges are uncertainties about what to test and how to test, lack of defining test strategies and not setting up optimal test environments, causing after effects in the form of insufficient data in the production environment, thus, resulting in deferment in implementation, which causes an increase in the cost [9]. Therefore, testing becomes a critical concern during different phases of the Big Data pipeline [10]. However, testing of Big Data also challenges the traditional testing techniques. Considering the importance of Big Data testing, we decided to conduct a systematic literature review to determine what testing techniques have been used to test the Big Data pipeline and how Big Data testing is challenging traditional testing approaches.

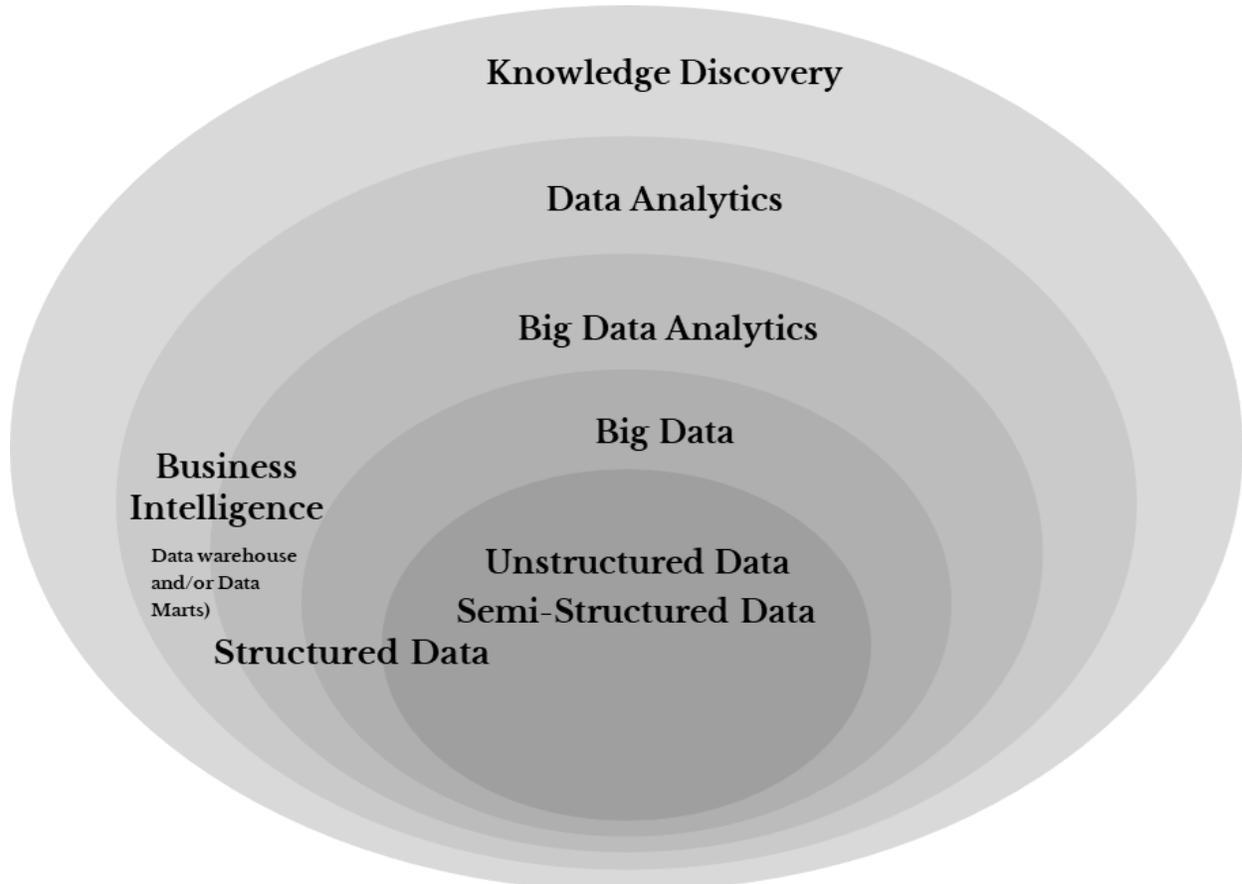

**Figure 1:** Visual Representation of Big data, Big Data Analytics and Business Intelligence (adapted from [5]).

### *1.1 Related Work*

Big Data is a series of approaches, tools and methods for processing a high volume of structured, semi-structured or unstructured data. Therefore, it is hard for testers to test and validate the large scale of data and features. Many existing studies discuss data quality issues along with data characteristics. For instance, the authors in [11] performed a systematic mapping study by covering software testing in the context of MapReduce programs. The results reveal that the most frequent reason for testing MapReduce programs is performance issues and improper use of the processing model. In addition, most testing efforts are about the performance and less focus on testing the functional aspects of MapReduce.

In [12], the authors addressed the current state of challenges for Big Data applications in terms of data quality. Similarly, the authors of the study in [13] discussed test data management in a data warehouse and performance testing. In [14], the authors proposed the idea of a new test factory model and the quality management system for Big Data applications. While briefly reviewing the literature for test techniques for big data, they mainly emphasize an end-end quality management model for Big Data applications. Similarly, in [15], the authors discuss general quality assurance techniques for Big Data applications such as testing, model-driven architecture, monitoring, fault tolerance, verification and prediction. In [16], the authors reviewed two critical aspects of Big Data bioinformatics analysis: scalability and validity of the software. They also discussed software testing techniques based on the idea of multiple executions, such as metamorphic testing. They mentioned that these techniques could be used to implement an effective bioinformatics quality assurance strategy. Furthermore, the authors in [17] discussed the benchmarking of Big Data systems. They provided a comprehensive view of benchmarking by including Big Data micro, end-to-end and benchmark suite.

In [18], the authors reviewed the recent progress and breakthroughs of big data applications specifically in healthcare domains and summarized the challenges, gaps and opportunities to improve the advanced big data applications in health care. In [19], the authors conducted a mapping study of testing MapReduce (MR) focusing on tool, environment, testing and faults. The research results show that concerning MR program testing, there were gaps to be filled and challenges to be overcome. In another study [20], the authors attempted to highlight Big Data testing but nothing related to testing techniques and challenges has been done.

In short, there is no existing literature review discussing Big Data testing techniques and challenges. To the best of the authors' knowledge, this paper is the first literature study available that gives a systematic conclusion on Big Data testing techniques and associated challenges in terms of scope and coverage of multiple literature sources. Tab. 1 compares the existing work related to Big Data testing.

**Table 1:** Existing Literature and our Contribution.

| Number | Highlights |
|--------|-----------|
| [11] (2019) | Covering software testing in the context of MapReduce programs. |
| [12] (2016) | Addressing the challenges and Big Data applications in terms of data quality. |
| [13] (2013) | Test data management in a data warehouse and performance testing. |
| [15] (2015) | Quality assurance techniques for Big Data applications including testing, model-driven architecture, monitoring, fault tolerance, verification and prediction. |
| [16] (2017) | Multiple execution metamorphic testing for validation of Big Data bioinformatics software. |
| [17] (2017) | Benchmarking of Big Data systems. |
| [18] (2016) | Big Data applications in biomedical health care. |
| [19] (2013) | Big Data mapping study for testing MapReduce. |
| [20] (2021) | Big data testing without discussing the testing techniques and challenges. |
| OUR | Covers diverse testing techniques used to cope with different challenges of Big Data with, covering functional and non-functional testing techniques and the challenges faced while testing Big Data. |

## *1.2 Motivation and Contributions*

Big Data testing has several benefits: improving business decisions, quality, reducing losses, improving market strategies, and improving revenues. In Big Data Systems, Big Data testing is critical to assure their proper functioning. If these systems are not adequately tested, companies are bound to be negatively impacted due to potentially wrong and untimely decisions. Furthermore, it will be challenging to figure out what went wrong, what caused the failure and where it happened. As a result, finding a solution to a problem will be challenging.

On the other hand, if Big Data testing is done effectively, it will help avoid resource waste. This paper attempts to fill the gap in reviewing available evidence on Big Data testing. First, the few informal and formal review studies that exist [11-20] in the area deal with issues such as data quality, challenges in testing Big Data applications, overall quality assurance of Big Data applications with testing being only a sub-part of it, benchmarking of Big Data systems and testing of Big Data in specialized domains such as bioinformatics and MapReduce testing. Secondly, no systematic literature review summarizes the testing techniques along with the associated challenges for Big Data by covering multiple literature sources. The contributions of this paper are as follows.

1. We discuss and highlight Big Data testing techniques and uses. Then, we classify diverse techniques into functional, non-functional and combined (functional and non-functional) categories.

2. We propose a conceptual pipeline for Big Data testing processing.

3. We classify Big Data testing challenges and then identify proposed solutions based on one testing technique.

The remaining part of this paper is organized as follows. Section 2 explains the systematic review process of this study. Section 3 presents the testing techniques to test Big Data. Section 4 discusses the Big Data pipeline and phases. Section 5 discusses the challenges and future trends, Section 6 presents a discussion of the paper results along with the validity of this study. Finally, Section 7 concludes the paper.

## 2 Review Process and Analysis

Systematic Literature Review (SLR), also known as a secondary study, is primarily a process to gather evidence from primary studies based on a focused topic. SLR is conducted according to defined, systematic methodological steps (called as the review protocol). To achieve this, we have followed guidelines of conducting a systematic literature review in software engineering [21-23]. Our resulting review protocol has some phases, i.e., planning the review, conducting the review and documenting the review. In planning the review, we define our research questions and define the search strategy. While conducting the review, we define our study selection process and data extraction procedures. In the end, the documented results of the review are reported. The described phases are not followed in a sequence; it is an iterative process based on continuous feedback from each phase.

## *2.1 Planning the Review*

In the first phase of this review, we specified the research questions and defined the search strategy.

### *Research Questions (RQs)*

In order to gather the evidence regarding software testing techniques for Big Data applications along with associated challenges, we formulated the following research questions:

RQ1: What are the existing techniques for testing Big Data?

RQ2: What type of challenges or limitations exist in testing Big Data?

The research questions are planned and structured using the recommended PICOC (Population, Intervention, Comparison, Outcome, Context) criteria [22]. However, we do not impose any restrictions in terms of comparison and context. The population in this study is the domain of Big Data applications. *Intervention* is represented by the techniques to test Big Data application(s). The outcome of our study represents a comprehensive collection of different types of test techniques and challenges to testing Big Data.

### *2.2 Generation of a Search Strategy*

The main objective behind the search strategy was to identify an exhaustive and unbiased set of primary studies to answer the defined research questions. For this purpose, the search strategy was broken down into distinctive steps: i) Identification of search terms, abbreviations and synonyms to support the research questions. We finalized the following main search term: Big Data software testing. ii) Use of Boolean OR to join different terms. iii) Use of Boolean AND to join major terms. The search is limited to keywords and/or index terms, titles and abstracts in each electronic database. We started defining the scope and the search terms during May 2021 and ended in the second week of May 2021, leading to the following search string: ("Document Title": Big Data software testing) OR "Index Terms": Big Data software testing) OR "Abstract": Big Data software testing). We searched the papers in major electronic databases for the past ten years (2010-2021). The search string used for each database is given in Tab. 2.

**Table 2:** Verbatim Search Strings as Executed in Electronic Databases.

| Source | Search String |
|---|---|
| IEEEXplore | (((''Document Title'':Big Data software testing) OR ''Index Terms'':Big Data software testing) OR ''Abstract'':Big Data software testing |
| ScienceDirect | Title, abstract, keywords: Big Data software testing |
| Scopus | TITLE-ABS-KEY (big AND data AND software AND testing) AND (LIMIT-TO (SUBJAREA, ''COMP'') OR LIMIT-TO (SUBJAREA, ''ENGI'') OR LIMIT-TO (SUBJAREA, ''MATH'') OR LIMIT-TO (SUBJAREA, ''DECI'')) |
| ACM | Abstract:(+big +data +software +testing) AND acm dl Title:(Big Data software testing) |
| SpringerLink | 'big AND data AND software AND testing AND (big OR data OR software OR testing)') |

Endnote X9.3 (a bibliography management tool) was used to record the results for each electronic database and to remove duplicate references. We set time/year restrictions between (2010 and 2021), while executing the automated search string. Using the search strings mentioned in Tab. 2 resulted in a total of 1975 studies. The first step was to remove the duplicates using the Endnote X9.3 tool as the same papers were indexed in multiple databases. After duplicate removal, we were left with 1838 studies. The distribution of the original count per data source distributed through the Endnote X9.3 tool is given in Tab. 3.

**Table 3:** Distribution of Papers among Different Sources.

| Source | Count (before duplicate removal) |
|---|---|
| IEEEXplore | 280 |
| ScienceDirect | 105 |
| Scopus | 771 |
| ACM | 755 |
| Springer | 64 |
| Total | 1975 |

### 2.3 Conducting the Review

There are several steps involved in conducting the review: study selection criteria and procedures for including and excluding studies (Section 2.3.1) and data extraction (Section 2.3.2). This Section describes these steps, while Fig. 2 illustrates the selection process and the number of primary studies identified at each stage.

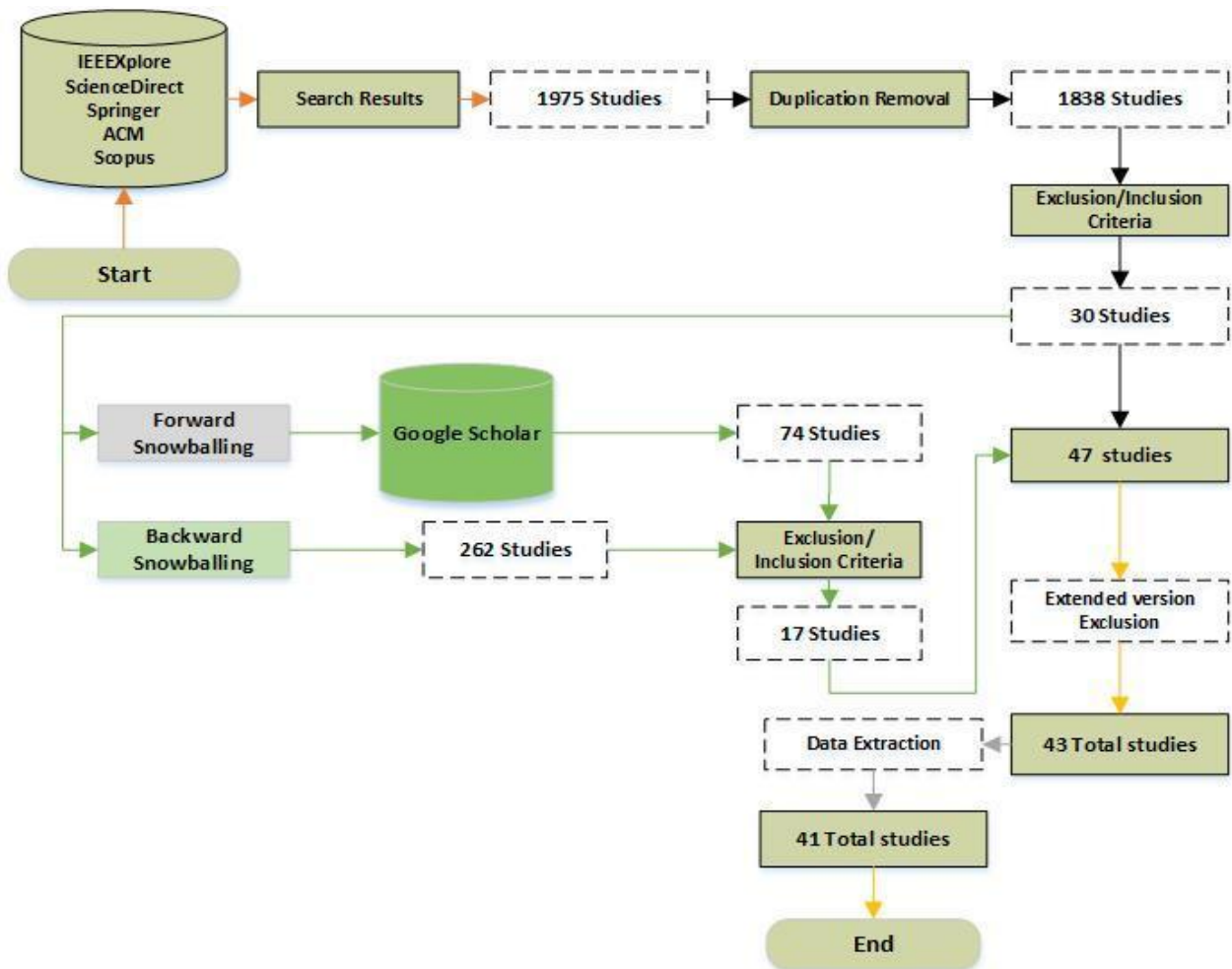

**Figure 2:** Overview of the Selection Process and Final Number of Primary Studies.

### 2.3.1 Study Selection Criteria and Procedures for Including and Excluding Studies

Study selection criteria are used to identify studies that provide direct evidence for answering the stated research questions presented in Section 2.1. Tab. 4 shows the inclusion (I) and exclusion (E) criteria applicable in this review. Our study selection criteria are based on multiple steps. We created groups of papers per publication database in Endnote to maintain the references count after duplication removal. For each database, we created five sub-groups: rejected papers, final accepted, book and thesis, conference venue and details, review papers and in the end, not clear papers. The purpose of this grouping was to reach a logical categorization of references.

**Table 4:** Inclusion and Exclusion Criteria.

| Exclusion Criteria | |
|---|---|
| E1 | Papers that are not related to Big Data testing. |
| E2 | Papers that merely mention Big Data testing but it is not the focus. |
| E3 | Papers whose titles and abstracts contain no information about Big Data and testing. |
| E4 | Papers that are simply a foreword to conference proceedings. |
| E5 | Papers that are published prior to 2010. |
| E6 | Books, thesis, review papers. |
| E7 | Papers that are written as a precursor to presentations and newsletters. |
| E8 | Papers that are related to cloud computing and data quality. |
| E9 | Papers related to Big Data but do not contain any information related to our research questions. |
| E10 | Papers that are not in English. |
| **Inclusion Criteria** | |
| I1 | Papers related to testing techniques for Big Data applications. |
| I2 | Papers containing challenges for testing Big Data applications. |
| I3 | Papers coming from acceptable, peer-reviewed sources: workshops, conferences and Journals. |
| I4 | Papers containing any information related to our defined research questions. |

For instance, while quickly scanning the references, the first researcher placed several references in the review papers, conference venue and details, book and thesis groups per database. Furthermore, based on reviewing the titles and abstracts (according to the inclusion and exclusion criteria), clearly out of scope references were placed in the rejected papers group. In contrast, accepted papers went into the final accepted group. Papers in the not clear papers group required skimming the full-text, which eventually placed them in either rejected papers or final accepted papers groups. The counts' distribution at each stage of this filtering process is further elaborated in the steps described below. In the beginning, a single researcher was involved in this activity. However, while quickly reviewing the papers based on their titles and abstracts, it emerged that 67 studies were merely describing the conference venues and details, 22 were books and theses, while 7 were review papers. Excluding these references, we were left with 1742 out of the initial 1838 (duplication removed) references. Based on the titles and abstracts, 1589 out of 1742 papers were excluded because they were clearly out of the scope and unrelated to our proposed research questions. At the end of this activity, 153 papers were left for further assessment.

Out of the 153 studies, 123 were further rejected by critically analyzing the papers by the authors, with 30 papers left behind. Furthermore, to make sure we do not miss any relevant paper, we applied forward and backwards snowballing [24, 25] on the references of these 30 papers, resulting in a total of 336 more papers as the outcome. We applied our inclusion/exclusion criteria on these 336 papers and were left with 17 more to add to the previous 30 papers, making the total number of papers equal to 47. Finally, the 47 papers were analyzed thoroughly to determine whether it is possible to extract the data required to answer our stated RQs. During this analysis, we found that some papers were extended versions of previously published conference/workshop papers [26-29]. Moreover, we were unable to extract data from two papers [30, 31]. Therefore, after removing these papers, 41 papers were finally selected as primary studies for answering the RQs in this SLR.

### 2.3.2 Data Extraction

To extract the relevant data from the final pool of 41 primary studies, a spreadsheet-based data extraction form was created. An existing template [32] was followed as an example. The form was designed so that the extracted data would address our research questions. Specifically, we extracted data for two categories: paper information and topic-specific classification. The data extraction form and the mapping of research questions are shown in Tab. 5.

**Table 5:** The SLR Data Extraction Form.

| Data Item | Value | RQ |
|---|---|---|
| **Paper Information** | | |
| Primary Study (PS) ID | Integer | - |
| Paper Title | Name of the paper | - |
| PDF Link | Link to the paper | - |
| Abstract | Abstract of the paper | - |
| **Topic Specific Classification** | | |
| Challenges | Big Data testing challenges | RQ 2 |
| Testing techniques | Techniques list | RQ 1 |
| Testing Types | Info. about types | RQ 1 |

## 3 Big Data Testing Techniques Analysis and Synthesis

This Section presents the results of this SLR, answering our stated RQs.

### 3.1 Testing Techniques (RQ1)

This Section presents the software testing techniques to test Big Data applications that we gathered from the selected primary studies. Fig. 3 presents the classification of each testing technique into functional, non-functional and combined (functional and non-functional). Lastly, we also present testing techniques and papers in Tab. 6.

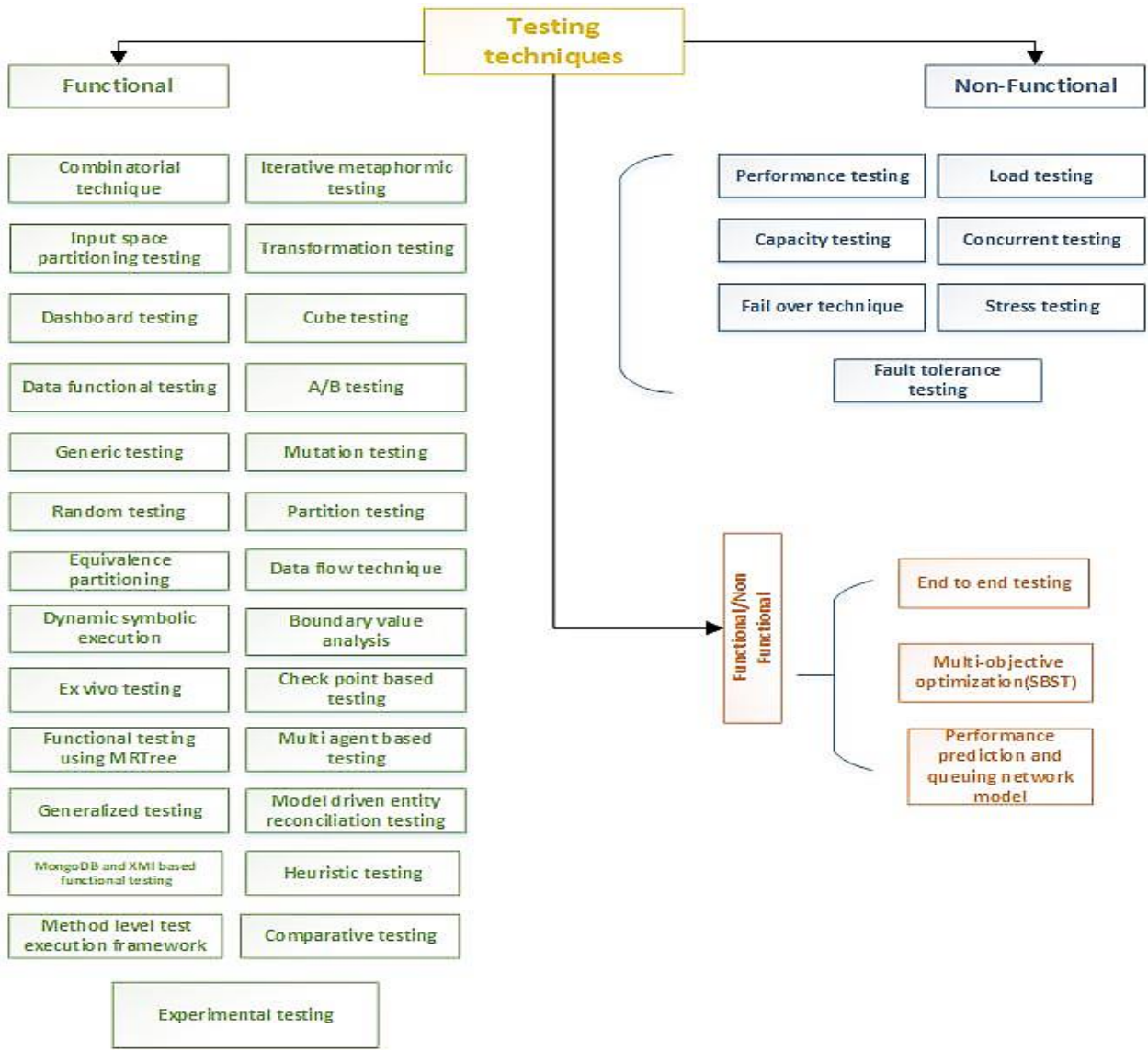

**Figure 3:** Identified Testing Techniques for Big Data Testing.

**Table 6:** Test Techniques Identified in Primary Studies.

| Type | Testing Technique(s) | Primary Studies |
|---|---|---|
| | Cube testing | [9] |
| | Dashboard testing | [9] |
| | Data functional testing | [9] |
| | Performance testing (**NF**) | [9] |
| | Failover testing (**NF**) | [9] |
| | Combinatorial technique | [33, 34, 35] |
| | Generic testing | [35] |
| | Combinatorial technique | [36] |
| | Random testing | [36] |
| | Equivalence partitioning testing | [36] |
| | Data flow testing technique | [37] |
| | Ex vivo testing | [38] |
| | Functional testing using MRTree | [39] |
| | Dynamic symbolic execution | [40] |
| | A MongoDB and XML-Based Functional Testing Technique | [41] |
| | Transformation testing | [42] |
| | Partition testing | [43] |
| | Combinatorial technique | [43] |
| | Random testing | [43] |
| | Mutation testing | [43] |
| | Iterative metamorphic testing | [43] |
| | Model driven entity reconciliation testing | [44] |
| | Check point based testing | [45] |
| | Input space partitioning | [46] |
| | Equivalence partitioning testing | [47] |
| | Boundary value analysis | [47] |
| | Load testing (**NF**) | [47] |
| | Capacity testing (**NF**) | [47] |
| | Concurrent testing (**NF**) | [47] |
| | Stress testing (**NF**) | [47] |
| | Performance testing (**NF**) | [47, 48] |
| | End to end testing (**F and NF**) | [48] |
| | A/B testing | [49] |
| | Data functional testing | [51] |
| | Failover testing (**NF**) | [51] |
| | Multi-agent based approach testing | [52] |
| | Generalized testing | [53] |

| | |
|---|---|
| Method level test execution framework | [54] |
| Multi-objective optimization (SBST) (**F and NF**) | [54] |
| Experimental testing | [55] |
| Comparative testing | [56] |
| Heuristic testing | [57] |
| Fault tolerance testing (**NR**) | [58] |
| Performance prediction and queueing network models (**F and NF**) | [59] |

### *3.2 Combinatorial Techniques*

Li et al. [33] have used combinatorial test data generation technique to test Extract, Transform, Load (ETL) applications and have named their tool combinatorial Big Data Test data Generator (BIT-TAG). They aim to generate small, efficient datasets. While using manual Input Domain Models (IDMs), problems occur due to changes in the original data source or constraints. IDMs are created from an original data source based on input parameters and test values; these parameters and tested values have been derived from some constraints, either defined by clients or extracted by database schema. Therefore, whenever changes occur in the original data source, re-processing in IDMs is needed too. New IDMs, referred to as Adaptive IDMs (AIDMs), have been proposed to handle this problem. These adaptive IDMs are created by the BIT-TAG tool after analyzing the extracted test values from the original data; the generated datasets satisfied t-way combinatorial coverage and other constraints.

Fredericks and Hariri [34] have proposed search-based software testing and combinatorial testing to test an unmanageable set of operational contexts and configurations for a Big Data application. As an application, the authors [34] gave an example of one Big Data source: a Medical Records Network (MRN). Suppose an application interface with an MRN to retrieve data. In that case, the test generation can be optimized for a specific type of data, such as Binary Large Object (BLOB) data (i.e., images, scans). Different test suites could be combined to enhance combinatorial coverage to cover combinations of configurations.

Moran et al. [35] have argued that wrong infrastructure configurations on MapReduce applications affect the program output, causing functional problems which are very difficult to expose while testing. Their approach uses combinatorial testing to generate different automatic configurations (also called scenarios) for MapReduce applications. Furthermore, the test execution engine based on the MRUnit library is used to automate the proposed approach. The proposed approach works based on a mapper, combiner and reducer where MRUnit executes each scenario. The mapper, combiner and reducer can interpret different data on each execution due to MapReduce programs' distribution environment. A combinatorial technique (i.e., combining values of different parameters) generates each scenario for different distributed environment configurations.

In a study [36], Moran et al. proposed two new test techniques for MapReduce applications. One is called MRTest-Random based on random testing and the second is called MRTest-t-Wise, which is based on equivalence partitioning with combinatorial testing (a kind of partition testing). Given the configurations to test, MRTest-Random randomly generates valid configurations. Next, these configurations are divided into partitions using MRTest-t-Wise based on similarity of behaviour. A combinatorial strategy is then applied to generate test configurations automatically. Finally, MRTest (an execution engine) runs each test configuration and systematically checks if all the configurations lead to equivalent outputs.

Ding et al. [43] have used combinatorial techniques combined with random testing and a categorical choice framework to produce initial test inputs. In addition, it is challenging for a 3D reconstruction program to check the consistency between a reconstructed and actual cell. For instance, for a given input to ADDA (i.e., light scattering simulator based on the discrete dipole approximation), it is also hard to know the correctness of the output. Moreover, these are typically non-testable due to the absence of test oracles. Therefore, these techniques helped them to test non-testable oracles.

### 3.3 Combination of Two Techniques

In [36], Moran et al. proposed two new test techniques for MapReduce applications. One is called MRTest-Random based on random testing, and the second is called MRTest-t-Wise, which is based on equivalence partitioning with combinatorial testing (a kind of partition testing). First, given the configurations to test, MRTest-Random randomly generates valid configurations. Next, these configurations are divided into partitions using MRTest-t-Wise based on similarity of behavior. A combinatorial strategy is then applied to generate test configurations automatically. Finally, MRTest (an execution engine) runs each test configuration and systematically checks if all the configurations lead to equivalent outputs.

### 3.4 Data Flow-Based Testing Technique

Moran et al. [37] proposed a data flow test criteria-based testing technique called MRFlow (MapReduce Data Flow) to detect MapReduce programs' defects. MRFlow analyses the evolution of variables in the MapReduce programs. A MapReduce program functionality is represented utilizing program transformations to deal with heterogeneous data sources and formats. The test cases are derived from such transformations based on extracting the paths under test from the program graph and then testing each path under test with different data.

### 3.5 Ex Vivo Testing

Moran et al. [38] have proposed an Ex Vivo testing framework called MapReduce Ex vivo testing (MrExist) that takes advantage of the production information to detect faults. The authors identified some faults in their previous work [39]. Therefore, they proposed an automatic testing framework to detect the faults when deployed and executed in production. These faults may depend on the deployed MapReduce configuration. Therefore, one would need to test the application in production (In Vivo). This testing is hardly feasible due to a lack of control on the tester's side. Therefore, the authors propose a hybrid approach between testing in the laboratory and testing in production. They have named the Ex Vivo approach: the tests are automatically obtained from the runtime data but executed outside of the production environment to not affect the application. The proposed framework can automatically detect functional faults without requiring any human intervention.

### 3.6 Functional Testing using MapReduce Tree (MRTree)

In [39], Moran et al. proposed a hierarchical fault classification-based approach called MapReduce Tree (MRTree) that identifies functional faults in MapReduce programs. Test cases are derived to detect the faults represented by MRTree nodes. An example program as the system under test for each node failure is described for each node failure and guidelines to produce test cases.

### 3.7 Dynamic Symbolic Execution

Csallner et al. [40] have used a dynamic symbolic execution method for testing MapReduce programs where certain correctness conditions need to be met. However, if these conditions are not met and if there are variations in the infrastructure events such as network latency, scheduling decisions and so on, the output of MapReduce programs can be erroneous. This can lead to MapReduce programs giving variable outputs on the same input data. The technique encodes the high-level MapReduce correctness conditions as symbolic program constraints and checks them for the program. A constraint solver is used to infer program input values that violate the correctness conditions.

### 3.8 A MongoDB and XML-Based Functional Testing Technique

Yesudas et al. [41] proposed a Big Data performance test framework based on a NoSQL database for a highly transnational back-end system like the IBM sterling order management system. They created a proof of concept prototype based on distributed storage to ensure that the system performs to meet the retailers' peak hours when the massive orders are received. XML formats are used because these are 10% faster than the traditional performance tools such as LoadRunner and Grinder. MongoDB query is used for handling large files of any JSON and real-time data. The proposed framework is validated by creating XML messages in MongoDB collections (i.e., generation of load) and the initial data flow is started by running the scripts.

### 3.9 Transformation Testing

Tesfagiorgish and Yi [42] proposed an approach to test the correct transformation of huge volumes of data. Their approach is based on reverse engineering, where the whole transformation process is reverse engineered to retrieve the source. Then, the output (target data) of the ETL operation is given as an input to a reverser, which expects its output (the reversed source data) to be the same as the source data.

### 3.10 Iterative Metamorphic Testing

Ding et al. [43] proposed an iterative metamorphic testing technique to test an online Big Data service that includes scientific tools, machine learning algorithms and an extensive scale image database. The technique consists of three significant steps: 1. development of initial Metamorphic Relations (MRs) and tests (the initial inputs were developed and converted into another set of inputs as per MR); 2. test evaluation with program coverage criteria and mutation testing (where mutated tests were tested, whose output violated an MR. Every mutant was killed when the output of the mutant program deviated from the original one); 3. refinement of MRs (was done by analysing the data using advanced machine learning techniques such as Support Vector Machine (SVM).

### 3.11 Model Driven Entity Reconciliation Testing

Blanco et al. [44] presented a model-driven engineering approach for integration testing of Entity Reconciliation (ER) applications. ER applications combine data from different sources to present a unified view of some real-world entity. In this approach, the authors have built the Integration Testing Rules (ITR) model based on a reconciled solution model, data sources model, the transformations model and the test models. The testing objectives are represented as business rules covered by applying the masking modified condition decision coverage criterion.

### 3.12 Checkpoint-Based Testing

Sudsee and Kaewkasi [45] proposed a Distributed Test Checkpointing (DTC) method for unit testing of Big Data applications developed using the Apache Spark framework. Spark framework of data-intensive processing is implemented in the form of a Resilient Distributed Dataset (RDD). DTC identifies each data partition of a RDD and the test case that uses the RDD using a hash function. Using this, DTC speeds up unit testing on a distributed cluster repeatedly.

### 3.13 Input Space Partitioning Testing

Li et al. [46] proposed a scalable novel Big Data testing framework for solving technical problems faced during ETL testing. These technical problems are long processing time due to massive data transformation among different data sources and the need to validate the transformed data. The authors have used the input space partitioning technique to generate test data. They control the test set's size and generate a small, efficient test set by using Input Domain Models (IDMs). To select and generate tests, they partitioned

IDMs, selected the test values from each partitioned block, and generated tests by applying combinatorial coverage criteria. To validate transformed and transferred data, they have defined two rules. According to the first rule, they have extracted all the data types and value ranges from the requirements at a high level and, as per the second rule, extracted specification details to validate the transformation rules. Therefore, the authors of [46] have provided the solution for addressed problems, validated that transformation has been done correctly, and also ensured that the data is flawless by comparing source and target data.

### 3.14 Performance Testing Framework

Liu et al. [47] proposed a performance testing framework to test the Big Data applications. They highlight different performance test phases in the light of Big Data applications that need to be considered while testing performance tests. This framework includes test goal analysis, test design and load design for Big Data applications. During defining the goal phase to ensure the stability and reliability of big data systems, various test types need to be considered. For instance, concurrent testing: concurrent hit and access same module of the same application to check performance behavior, load testing: check the behavior of the system under a certain level of load to satisfy the user requests within the required range, stress testing: check the system behavior under some extreme conditions such as peak capacity, beyond the limit load examine the behavior of the system and capacity testing: perform to check the availability of a maximum number of resources without failure of the system. Whereas while defining test design, size of data and complexity of application should be in focus by using some manual techniques such as equivalence partitioning (i.e., includes the equivalence valid and invalid classes and based on specifications design test data) and boundary value analysis (i.e., design the boundary conditions for test data).

While load design, indicators need to be provided for an in-depth system performance analysis. These test indicators include response time (i.e., time to respond to a request in addition with server response time), throughput (i.e., number of requests processed per unit time) and monitoring indicators for monitoring resources like CPU memory utilization, storage, etc.

Nachiyappan and Justus [48] discussed that Big Data applications lead to the worst situations due to erroneous architecture elements while designing them, thus resulting in real-time scalability issues. Therefore, performance testing needs to skip down the scalable issues that have been faced while testing.

Gudipati et al. [9] also discussed the need for non-functional ( i.e., performance testing) techniques for Big Data applications. They have addressed that the performance of a system went down due to various reasons, such as enormous data size, negligence in designing architecture and the need to process data across multiple nodes to complete the job at a convenient time. Therefore, performance testing should be performed to eliminate the aforementioned performance issues. This testing can be performed using some Hadoop performance (i.e., Job completion time, throughput) and system-level (i.e., memory utilization) metrics to identify bottlenecks and ensure the smooth performance of such applications..

### 3.15 End to End Testing

Nachiyappan and Justus [48] discussed that dealing with the huge amount of data and executing it on multiple nodes can create bad data and quality issues at every stage of processing. Therefore, there is a need for integration data available all the time to ensure quality. At this stage, the end-to-end testing technique ensures that flawless and stable data should be available among all data sources.

### 3.16 A/B Testing

Liu et al. [49] discussed online experiments considering Big Data to provide business guidance based on some analytic data techniques; experiments have been conducted to understand customers' behaviour. However, there is no proper guidance on data sample size to have valuable, trust-able discovery to mine market information. A/B testing (i.e., bucket or split testing) is commonly used for meaningful marketing and business intelligence conclusions. Therefore, [49] conducted an online experiment using A/B testing. Finally, the author concluded that if we had insufficient or not enough data samples in the control experiment, results could not be completed even after investing a long time.

### 3.17 Functional Testing by using HadoopTest (for Test Harness)

Marynowski et al. [50] discussed that MapReduce implementations on a large-scale distributed environment fails due to several issues such as hardware problems, outages and bugs. Therefore, implementations should be designed to be fault-tolerant; and tested in such a way to make sure reliability. They have proposed a HadoopTest framework for MapReduce based systems to encounter this problem. This framework has put together functional and fault injection tests to build complex, complicated test cases. On the other hand, for source code, API calls have not been added to the system under test cases to avoid adding more bugs, making it more difficult to detect the source failure.

### 3.18 Failover Testing, Cube and Dashboard Testing

Garg et al. [51] and Gudipati et al. [9] mentioned functional data testing for Hadoop architecture, called failover testing. The Hadoop architecture consists of a name node and other connected data nodes hosted on several servers. So some of the failures that can happen are named node failures, data node failures and network failures. The main purpose of failover testing is to ensure the verification and validation of recovered data when switching from one node to another after failure. Further, Gudipati et al. [9] discussed the validation of the analytical reports. When ETL is processed and data is loaded into enterprise data warehouse tables successfully, reports have been generated using some reporting tools or queries on Apache Hive. Cube testing and dashboard testing are techniques to verify these analytical reports. Cube testing needs to verify that values are correctly displayed in the reports whereas dashboard testing ensures that all modules are correctly rendered and validate the fetched data from various web modules against the database.

### 3.19 Data Functional Testing

Garg et al. [51] discussed that during the functional testing of the Hadoop framework, data testing is required at every stage to make sure that processed data is accurate and flawless. The main motivation to use this technique is to identify data-related issues due to node configuration failure or coding errors. Gudipati et al. [9] have also discussed data-related issues that occurred because of incorrect codes or node configurations. Thus, data testing should be performed on Hadoop validation phases to ensure the immaculate data is being processed.

### 3.20 Multi Agent Based Approach for Testing

Testing a distributed system such as Mapreduce is costly and time-consuming. Therefore, to efficiently test these systems. Hsaini et al. [52] incorporated the agents in the proposed approach. These agents are intelligent. They monitor the distributed system's behavior and execute any required testing if a defect is found or any misbehavior is identified. Furthermore, all agents in this approach are autonomous and work in parallel, which is essential in decreasing the time and the cost. The test process in the proposed approach is performed in two steps. Furthermore, the authors also described the behavior of a tester while describing the test steps. In the first step, the behavior of the tester is to read and split data, send the generated data to mappers, receive the output from mappers, and check time constraints. Whereas, in the second step, the behavior of the tester is to shuffle and sort the data received from the first tester, send the generated data to the set of reducers, check time constraints and return the final results. Whereas the procedure of the first algorithm is testing of mappers used in MapReduce implementation under test and the second is to test the reducer. In the end, the proposed prototype of this model is under experimentation. They selected the Madkit tool as a Multi-agents environment to implement our prototype testing.

### *3.21 Validation of Data by Generalized Testing Framework*

Sharma and Attar [53] proposed a generalized testing framework to validate complete data and make sure that data should be flawless in such a way from target to source that there should be less movement in data. For this purpose, they proposed a scalable framework and supported any type of relational database management system and different versions of HIVE as per the validation needed. They have provided the support of semi-structured data (i.e., JSON and XML format) validation. A user interface is provided to monitor the data so that it helps business analysts make decisions based on the data and provides some recommendations based on the business rules.

### *3.22 Method Level Test Execution Framework*

Feng et al. [54] proposed a method-level test execution framework to solve the Big Data applications debugging issues arising from a large volume of data. Their method consists of two steps. In the first step, by running applications in original datasets, they record all small number of method executions, later referred to as method level tests (i.e., Unit tests). They selected seven methods from four machine learning algorithms and implemented them in the weka tool using Java. The datasets have been selected based on size and execution time. These method level tests further recorded and evaluated the selected seven methods based on three coverage criteria (edge coverage, edge pair coverage and edge set coverage). In the second step, they reduce the size of the inputs using a binary reduction technique while preserving the same coverage achieved by the original method-level test. In addition, test effectiveness is done using a common mutation testing PiTest tool.

### *3.23 Experimental Testing*

Kuang et al. [55] focused on the problems of existing technologies that arise during the vulnerabilities discovery, such as fuzzing, symbolic execution and taint analysis, and identified that these problems have more or less relationship with data processing functions. Therefore, they proposed a data processing function identification using a deep neural network to solve these problems. Furthermore, for the evaluation of the proposed method, they performed experimental testing to verify and validate the results of the proposed method.

### *3.24 Comparative Testing*

Krasilnikov and Putintsev [56] developed a system library to solve the large volume of data processing in the shortest possible time. In the proposed library, quick sort and insertion sort methods are used to organize array elements and several other ways. Comparative testing has been performed to check the effectiveness of the proposed library with different array volumes and the working machine configurations. The results show some advantages of the proposed sorting library as compared with today's popular existing solutions. For instance, with 10 million elements, the sorting process occurs 4.6 times faster with the proposed library. Furthermore, the proposed library sorts within 71.5 milliseconds, whereas the existing solution, such as Windows Thread, sorts within 331.7 milliseconds). Further increase of the array volume will also differ in speed between the library and the best of solutions in 4.6 times.

### *3.25 Heuristic Testing*

Al-Salim et al. [57] proposed a mixed-integer linear programming model by building processing nodes from source to data centers in the network that used Big Data to address the problem of power consumption. The heuristic testing approach ensures the energy-efficient solution's performance by following a software problem where packages are required for data processing on the network. The results were later compared with classical Big Data networks.

### *3.26 Fault Tolerance Testing*

The MapReduce system faced failure because of various factors such as network connections, interruption, outage, software defects, software updates and hardware issues. Therefore, it is crucial to ensure that failures should not cause a delay in the execution of MapReduce systems. Marynowski et al. [58] presented an approach based on the petri net reachability graph. For representing the fault cases in the model, MapReduce components act as dynamic items and could be easily inserted and removed without changing in the model. On the other hand, the HadoopTest framework automates the representative fault cases over the distributed environment. Moreover, monitoring and controlling each system component and according to their status injecting faults. This proposed method provides network reliability enhancements as a byproduct because it identifies errors caused by a service or system bug instead of simply assigning them to the network.

### *3.27 Performance Prediction and Queueing Network Models*

Zibitsker and Lupersolsky [59] have proposed the performance engineering process by using performance prediction and queuing network models to mitigate the performance risks. They have used test and production environment data for building the model of new applications and for models characterizing performance and resource consumption of the production workloads. Agents (i.e., Linux, Yarn, Kafka, Spark, Strom and Cassandra) have been used to collect data about resource information (i.e., CPU, memory, disk space usage and storage activity) and auto-detect agents are used to get information about node configurations (i.e., software and hardware configuration).

## 4 Big Data Pipeline and Testing Phases

In this Section, Big Data pipeline and the associated testing phases are discussed. Several phases are required, from collection to storing and refining the data during the Big Data pipeline. Fig. 4, illustrates a Big Data pipeline along with required testing phases. The details of these testing phases are discussed in Fig. 5. For example, in every phase of the Big Data pipeline, different testing techniques must be performed before moving to the next phase. The brief discussion on these testing phases is following:

**Data Validation:** In this phase, as the data is gathered from different sources (i.e., sensors, documents), there is a need to perform the necessary steps before transforming data to the next stage. Three key points need to be focused on these steps: i) Data gathered from multi forms, so there is a need for validation to ensure that correct data is pulled into the system, ii) Comparing source data with the target data into the Hadoop system and iii) Make sure that the correct data is extracted and loaded into the correct Hadoop Distributed File System (HDFS) location. During this phase, some of the testing techniques are the data flow-based testing technique and input space partitioning technique ([37] and [46]).

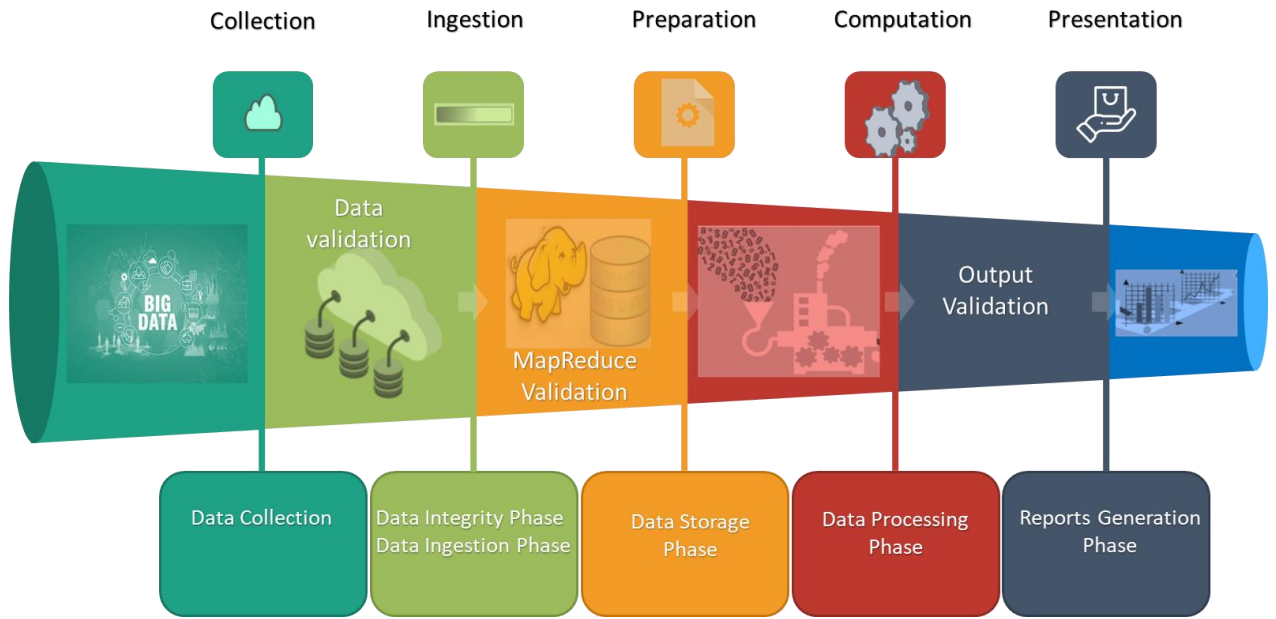

**Figure 4:** Big Data Pipeline along with Test Phases.

**MapReduce Validation:** In this stage, data transfers to any Big Data framework like Hadoop, Hive, Spark and the data is split between different nodes. In this stage, testing of data is divided into four categories: i) Ensure that nothing is lost in the data split process and that consistency should be in each node, ii) Data segregation rules are implemented on the data, iii) Expected MapReduce operation is performed and the Key-value pairs are generated and iv) Validating the data after the Map-Reduce process. In this phase, diverse testing techniques have been applied to perform MapReduce testing. For example, some of the techniques are combinatorial techniques [33], model driven entity reconciliation testing [44] and checkpoint based testing [45].

**Output Validation:** The final or third stage of Big Data testing is the output validation process which encompasses ETL testing and report testing. Finally, the output data files are generated and are ready to be transferred to an enterprise data warehouse or any other system based on the requirement. The following activities need to be a focus from the testing context: i) make sure that transformation rules are performed correctly, ii) check the data integrity and make sure that data correctly loads into the target system and iii) make sure no data is corrupted by comparing the target data with the HDFS file system data. Report testing is also required in this phase. The example techniques are cube testing and dashboard testing [50] used to perform validation of data.

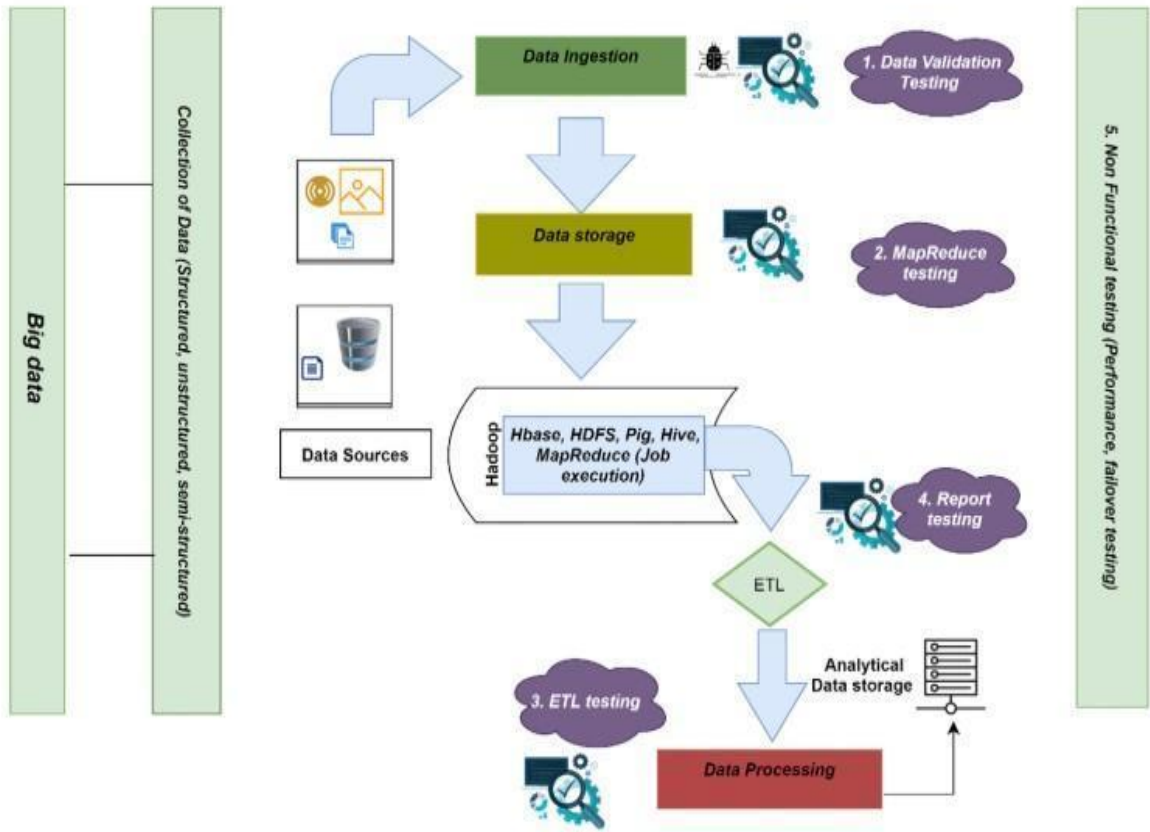

**Figure 5:** Testing Phases/Stages of Testing Big Data Pipeline.

## 5 Challenges and Future Trends

This section discusses the challenges of testing Big Data applications. In particular, we have mentioned the explored testing challenges in Tab. 7. Summarizing the testing challenges for Big Data (RQ2), we have observed that most of the testing challenges faced during ETL testing are due to the immense volume of data and discrepancy in transformed and transmitted data. Other challenges related to Big Data testing are performance, scalability, continuous availability and data security, meeting data speed and node failure. These challenges occurred for various reasons, such as the variable nature of data.

Furthermore, data security, the ability to handle larger workloads and distributed data environments, and chances of failure nodes (i.e., name, data and network) have increased, thus posing testing challenges. Other issues are related to general testing phases (i.e., requirement analysis and test case design, test data preparation, test environment and test management). In addition, validation of pre-Hadoop, report generation and data transformation also bring challenges. We have also observed a strong need for an automation testing framework to validate the data from source to target for the post ETL processes. Our findings show that most of the testing challenges occurred during the MapReduce validation (data processing and storing) phase. A lot of research is going on in proposing new testing frameworks to counter Big Data issues. However, all of these identified challenges during different testing phases need to be addressed to achieve quality in Big Data systems.

Moreover, there is also a need to focus on research in general testing phases for Big Data applications, such as test case design, test data preparation, test environment and test management. In future, we intend to enhance our research for Big Data Hadoop and microservices testing. A conceptual diagram related to challenges mapped with the phases of the Big Data testing is also illustrated in Fig. 6. A brief description of the mentioned above challenges are discussed below:

### 5.1 Data Processing Phase

Gudipati et al. [9] stated that Big Data testing is one of the biggest challenges faced by an organization dealing with Big Data. The critical factors are lack of knowledge about what to test and how to test, lack of defining test strategies and setting up optimal test environments, causing insufficient data in the production environment and associated after effects. In addition, Gudipati et al. highlighted numerous testing issues during the Big Data phases. In the validation of pre-Hadoop processing phase, data from various sources such as call logs, network sites and so on have been extracted based on defined requirements and loaded into HDFS Hadoop for further consideration. In this phase, the migrated data may be flawed for many reasons, such as incorrect data captured, incorrect storage of data, incomplete replication, etc. In another phase of validation of Hadoop MapReduce process, while MapReduce jobs run on the loaded data, plenty of issues occur like incorrect aggregations, node configurations, incorrect output format and incorrect code. These MapReduce jobs are streamlined while running on a single node, whereas, causes failure in case of processing on heterogeneous nodes. Furthermore, the authors have also discussed the issues faced during validation of reports such as report formats deviating from the requirements, data issues and formatting issues. Lastly, during validation of extracted data and loaded into an enterprise data warehouse, the issues include complete data extraction from HDFS Hadoop, incorrect loading of HDFS files into EDW and deviation in applying transformation rules.

### 5.2 ETL Testing Challenges

Yesudas et al. [41], Tesfagiorgish, JunYi [42] and Li et al. [46] highlighted challenges related to ETL testing. Yesudas et al. [41] highlighted testing challenges in data generation, transformation and loading of highly transactional systems like the IBM sterling order management system. However, Tesfagiorgish and JunYi [42], only focused on validation of transformation data challenge. According to them, validation of data transformation and identifying discrepancies in it challenges traditional testing techniques. Li et al. [46], discussed the three technical challenges that can be faced during the ETL testing. The first is related to generating a representative dataset due to the volume of variable data from variable resources such as embedded devices, clinical data, etc. The second is data validation and ensuring flawless data during load and transform is another big challenge. In the end, the biggest challenge is to validate an immense amount of data manually.

### 5.3 Data Quality and Node Failure

Nachiyappan and Justus [48] and Garg et al. [51] identified various challenges while testing Big Data applications:

**Performance:** Due to the variable nature of data from different sources, including different weblogs, sensors, embedded devices and so on, Big Data application performance testing challenges the traditional performance testing methods and techniques.

**Scalability:** The ability to handle the larger workloads and to accommodate the growing system is known as scalability. Currently, the dynamic growth of workloads is due to multiple factors such as business growth, new application features and usage patterns. Moreover, working with immense data in a distributed environment requires handling problems across multiple nodes. Therefore, data needs to be scaled rapidly across multiple data centers is a big challenge for Big Data applications.

**Continuous availability and data security:** Big Data contains sensitive information like personal ID, account, credit card details and so on. Therefore, the security of such information is very challenging due to massive data volume. Moreover, current No-SQL Big Data solutions have few mechanisms regarding data security.

**Meeting speed of data, understanding it and addressing data quality:** Big Data is challenging because it refines the data into an intelligence form to easily be used for data analysis and the target audience quickly consumes it. The inaccuracy of data affects the decision-making capabilities of organizations.

**Node failure:** Data is distributed across various nodes. Therefore, the chances of node failures such as name, data and network nodes increase, leading the organizations to significant losses. Testing for the prevention of these failures is challenging.

### *5.4 Post ETL Testing*

Sharma and Attar [53] highlighted some of the challenges of the post ETL process. They discussed the challenges regarding testing data in the data warehouse in terms of cost and the need for automation. They highlighted that manual validations of target and source data are very costly due to the massiveness of data. Moreover, in common practice, teams perform manual testing, which starts by comparing the data in an excel sheet with different means such as SQL scripting; copying the data from target and source databases is also very tedious and erroneous. Therefore, they suggested that an automated testing framework is required to validate data from target to source locally without moving out data from databases.

### *5.5 Unit Level Debugging Challenge*

Feng et al. [54] highlighted that debugging and testing Big Data applications is challenging at the module/unit level due to the large volume of data. They proposed a framework for effectively testing method-level tests to facilitate the debugging of Big Data. The proposed method consists of two steps. The first step was to generate a method-level test from a failed system-level execution while preserving code coverage. The second step was to reduce the size of the method level test by using the proposed binary reduction technique. This proposed framework helps developers to debug suspicious methods against the original input dataset while maintaining the high probability that the proposed method triggers failures caused by faults.

### *5.6 Vulnerabilities Challenges*

Kuang et al.[55] bring to light that data security is more concerned in the era of Big Data. In addition, they highlighted some of the problems with existing technologies used to discover vulnerabilities in the data processing software. Such technologies are fuzzy, symbolic execution and taint analysis. These have relationships with data processing functions. For example, in fuzzing, there are two types of sanity checks 1) critical and 2) non-critical, towards the target program. Bypassing such sanity checks leads to low coverage during fuzzing. Similarly, a constraint solver still has a problem dealing with complex algorithms' constraints in symbolic execution. Finally, in taint analysis, over-taint and under-taint affect the accuracy of results. To address these issues, they proposed a method named Data Processing Function Identification (DPFI) for identifying data process functions with the help of deep neural networks.

### *5.7 Data Gathering and Transmitting Challenges*

Al-Salim et al. [57] highlighted that efficient and economical transfer of data over the network in good time is a significant challenge. A large amount of processed data is either neglected, deleted or delayed. Furthermore, there is unnecessary power consumption, extra wastage of bandwidth and storage due to transferring raw data. The problems mentioned above result in increasing the financial and environmental costs. By considering these challenges, the authors proposed the mixed-integer linear programming model to study the impact of Big Data volume and variety on network power saving that carries Big Data traffic.

They employed the proposed technique to process Big Data raw traffic in the edge, intermediate and central processing stages. This has been done by building Processing Nodes (PNs) in the Internet Service Provider (ISP) network centers that host the Internet Provider (IP) over Wave-length Division Multiplexing (WDM) nodes. The volume scenarios captured generic results that demonstrate how the processing capability of the PNs dictates the Big Data volume that exists in Source Processing Nodes (SPNs), Intermediate Processing Nodes (IPNs) and Data Centers (DCs). They obtained up to 52% and 34% of network power savings in two different volume scenarios, compared to the power consumption of the classical processing approach where the Chunks are directly forwarded from the source node to the DCs.

### 5.8 Testing Challenges During Different Testing Phases

Zhou and Huang [60] and Staegemann et al. [61] focused on several common challenges while testing Big Data. However, Zhou and Huang [60], focused on testing one of the domains of Big Data, such as astronomical application. They highlighted the challenges in a different phase of testing. In the first phase, the Requirement analysis and test case design phase, many challenges are posed due to so many Big Data phases, while during data processing, defining scope is very challenging. Due to the characteristics of Big Data, test oracle is another big challenge. The diverse nature of data, such as structured, unstructured and semi-structured validation, is also a challenge. Moreover, the decision of actual and expected results is also a challenge. In the second phase, Test data preparation and test environment preparation of data for large scale testing data, generating test data during testing and lack of standardized data analysis language for the realistic, comparable workload is one of the significant challenges. In the last test management and the remainder phase, they discuss that testing Big Data challenges the traditional testing methods and results in increasing the cost of testing. In addition, the rule between deciding the actual results and the expected result is difficult due to the complexity of the software. Therefore, a tester should sound technical in different domains. Others bring into the notice challenges due to distributed development teams, part-time efforts, etc. Staegemann et al. [61] highlighted the Big Data testing challenges such as the oracle problem and additionally proposed guidelines for testing Big Data applications.

### 5.9 Performance Monitoring Challenges

According to Klein and Gorton [62], at the scaling of deployed systems and continually evolving workloads and unpredictable quality service of shared infrastructure, design time predictions are insufficient to ensure run time performance in production. Furthermore, due to the scale and heterogeneity of Big Data systems, significant challenges exist in the design, customization and observability capabilities. The challenges are efficient, economic creation of monitors to insert into hundreds or thousands of data nodes, low overhead collection and storage of measurements and application-aware aggregation and visualization. To counter these issues, the authors proposed a reference architecture that uses a model-driven engineering toolkit to generate architecture-aware monitors and application-specific visualizations to solve these challenges.

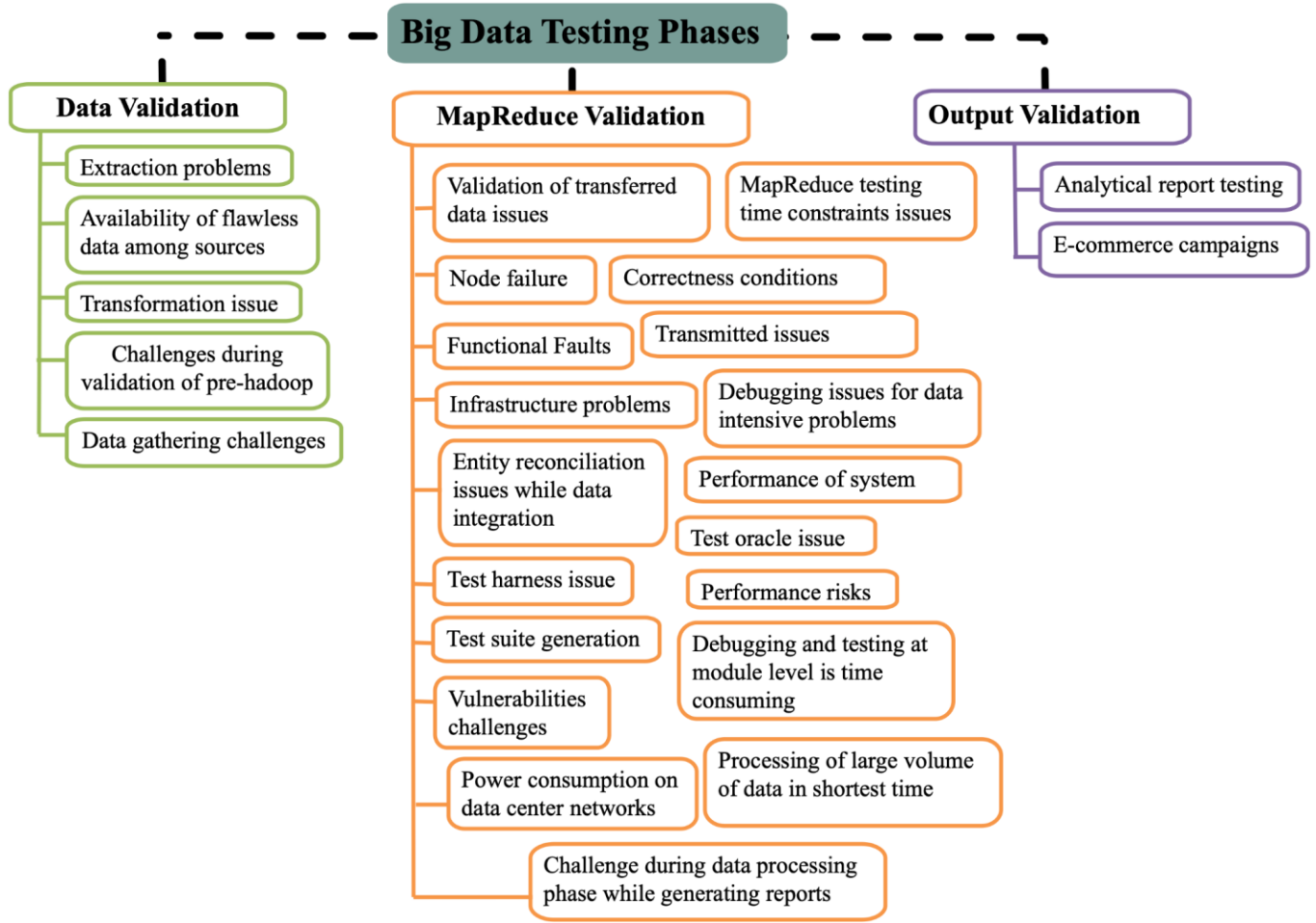

**Figure 6:** Conceptual Mapping of Identified Problems/Challenges with Big Data Testing Phases.

**Table 7:** Challenges Faced during Big Data Testing

| Testing challenges | Studies |
|---|---|
| Challenges during validation of pre-Hadoop, data processing phase and while generating reports | [9] |
| ETL testing problems | [41, 46, 53] |
| Transformation of data problems | [42] |
| Performance, scalability, continuous availability and data security, meeting speed of data, understanding it and addressing data quality and node failure | [48, 51] |
| Debugging and testing at module level is time consuming phases | [54] |
| Vulnerabilities challenges | [55] |
| Data gathering and transmitting challenges | [57] |
| Testing challenges during different testing phases | [60] |

## 6.  Analysis Discussion

In this Section, we will discuss our investigation on analyzing the software testing techniques for Big Data and challenges. Lastly, we also discuss the validity of our study.

Tab. 8 summarizes the challenges, their description and proposed solutions. The results show that the testing of Big Data systems would not only rely on traditional testing techniques but also on advanced techniques along with non-functional techniques (i.e., performance, scalability, availability, failover, end-to-end, stress, concurrent and fault tolerance). For instance, a most used testing technique to solve the Big Data functional faults faced during MapReduce validation phase is a combinatorial technique, in conjunction with other techniques such as random testing, mutation testing and input space and equivalence partitioning. The contributions of researchers in testing Big Data applications could not only help the testers in assuring the quality of data but could also help support the design of the test cases, automation tools to verify the data and monitoring tools to analyze the failures produced at run time in production.

**Table8:** Challenges during Big Data Processing and Proposed Solutions.

| Challenges | Description | Proposed solutions |
|---|---|---|
| ETL | 1. Generation of small yet efficient datasets to encounter the problem occuring during changes in original<br>data source or constraints.<br>2. Long processing time due to massive data, transformation among different data sources, need to validate the transformed data<br>3. Test the correct data transformation<br>4. Validate complete data and make sure that data is flawless from source to target<br>5. Dealing with the huge amount of data and executing it on multiple nodes can create bad data | 1. Combinatorial Big Data Test data Generator (BIT-TAG)<br>2. Scalable novel Big Data testing framework (Input space partitioning technique)<br>3. Transformation testing based on reverse engineering<br>4. Validation of data by generalized<br>testing framework<br>5. End-to-end testing |
| Test data generation | Unmanageable set of operational contexts and configurations for a Big Data application | Search-based software testing and combinatorial testing |
| Infrastructure configurations | Wrong infrastructure configurations on MapReduce applications affects the program output, causing functional problems which are very difficult<br>to expose while testing | Combinatorial testing along with MRUnit Library |
| Identification of design faults | Automatically testing design faults in MapReduce | MRTest-Random (Random testing),<br>MRTest-t-Wise (equivalence partitioning with combinatorial testing) |
| Data Flow testing | Detecting defects in MapReduce | MRFlow (MapReduce Data Flow) |
| Faults detection | 1. Automatic fault detection framework where tests are automatically obtained from the runtime data but executed outside of the production environment to not affect the application<br>2. Identification of Faults | 1. Ex Vivo testing framework called MrExist (MapReduce Ex vivo testing)<br>2. Fault Identification (MRTree)<br>3. Fault Tolerance Testing |
|  | 3. The MapReduce system faced failure because of various factors such as network connections, interruption, outage, defects, software updates and hardware issues. |  |
| Correctness conditions | Testing MapReduce where correctness conditions need to be met | Dynamic symbolic execution method for testing |
| Highly transactional back-end system | The system needs to meet the demands of retailers' peak hours when the massive orders are received. | A MongoDB and XML-based functional testing technique |
| Correctly transforming huge volume of data | Test the correct transformation of huge volumes of data | Transformation testing based on reverse engineering |
| Online Big Data Services | Big Data service that includes scientific tools, machine learning algorithms and a large scale image database | Iterative metamorphic testing technique |
| Entity Reconciliation (ER) applications | Entity Reconciliation (ER) applications combine data from different sources to present a unified view of some real-world entity | Model driven entity reconciliation testing |
| Unit testing of Big Data applications | Unit testing of Big Data applications developed using the Apache Spark framework | Checkpoint based testing |
| Performance testing Big Data application | A general framework for doing performance testing of Big Data applications | Performance testing framework |
| Analytical guidance | Business guidance for understanding the behavior of customers | A/B testing |
| Analytical reports | Validation of analytical reports | Cube and Dashboard testing |
| Functional testing | Functional testing for Hadoop architecture | Failover testing |
| Distributed systems | 1. Testing a distributed system such as the MapReduce system is a costly and a time-consuming process. 2. MapReduce implementations on a large-scale distributed environment fail due to several issues such as hardware problems, outages and bugs. 3. Power consumption | 1. Multi agent based approach for testing 2. Functional testing by using Hadoop Test (for Test Harness) 3. Mixed integer linear programming model testing |
| Performance risks | Performance risks and resource consumption of the production workloads | Performance Prediction and Queueing Network Models |
| Method level debugging | Debugging issues that arises due to large volume of data | Method level test execution framework |

| Data security | Problems of existing technologies that arises during the vulnerabilities discovering such as fuzzing, symbolic execution and taint analysis | Data Processing Function Identification using a deep neural network |
| Data processing | To solve the large volume of data processing in the shortest possible time | A custom build sorting system library |

### *Threats to Validity of the Results*

There are several threats to the validity of this review study. First, despite strictly following the guideline recommendations in [21-23], snowballing process guideline recommendations in [24, 25] and developing a review protocol for searching and selecting the primary studies (Sections 2.1 and 2.3), we generally accept that there might be missing primary studies due to one of the several reasons. We included all the papers that are in the English language, so there is a chance that we have missed out on relevant studies in other languages such as this paper [61]. We also believe this is the limitation of most of the SLR studies. We rejected all the studies that do not meet our inclusion criteria such as the papers [62-74]. Furthermore, we have decided to include all the primary studies that are only available electronically and are published. So there is a chance that relevant papers are not published due to privacy or other reasons. Unfortunately, our study does not aim to deal with such cases.

*Reliability validity:* Unbiased search strings have been tried in different electronic databases (Section 2.2) to assure the maximum coverage, though 100% coverage of relevant studies is not possible. To deal with this threat, we have performed an additional snowballing process defined in (Section 2.3.1). In addition, to minimize the risk of missing essential papers by using additional (snowballing) steps, we use Google scholar to scrutinize the citations. The main goal of this step is to include the maximum possible coverage of primary papers that might exist but missed by automatic search in Section 2.2. As an outcome of the search process, a total of 336 missing primary studies have been caught, out of which 17 studies meet our inclusion and exclusion criteria and have been selected for further analysis. Bias in including studies to answer our stated research questions is mentioned in Section 2.1. Some of the studies might be excluded due to our defined selection criteria (Section 2.3.1). We did not explicitly focus on MapReduce testing techniques as these studies can be found in [11, 19]. We only focused on including the papers that mentioned testing techniques to solve any Big Data related problems.

*Conclusion validity:* During the data extraction process, we have faced several difficulties with meeting our objective of extracting information, thus, resulting in a bias in primary studies. For instance, relevant terms are ambiguous, testing techniques not specifically used to address any specific Big Data problem, challenges not related to testing Big Data applications, etc. To minimize this bias, we first followed the well-defined data extraction form (Section 2.3.2) with the consent of the second author in order to ensure precise, relevant data extraction. Secondly, the first author extracts the data and the second author validates it to avoid any discrepancy in data. Finally, we believe that our review protocol is well-defined and detailed enough to evaluate how we select the final studies for further analysis and data synthesis.

*External validity:* Our study intended to be representative of included primary studies. We set no time restriction while executing the defined search string discussed in Section 2.2. This step aims to cover the maximum coverage of papers from the beginning related to Big Data testing. Finally, we select all the references for further analysis and consideration in the period between 2010 to 2021. We believe that our well-defined review protocol helps us achieve the representativeness of the final set of primary studies to a greater extent.

### 7.  Conclusion

This paper presents the exhaustive state-of-art of research in Big Data testing. Our research aims to systematically draw conclusions about the existing testing techniques used, summarize the testing challenges and highlight the research trends for Big Data testing. To achieve our goals, we have identified 1838 (duplication removal count) studies through bibliographic search; combining this with an additional snowballing (backwards, forward) process after inclusion and exclusion criteria, we ended up with 47 studies. Additionally, 41 studies were selected for further analysis and consideration based on our well-defined study selection process. We observed that the combinatorial testing techniques are used to solve particular Big Data problems combined with other techniques, i.e., random testing, input space partitioning, mutation testing and equivalence testing. Research in Big Data testing has been a primary focus area for practitioners and researchers in recent years. We found that traditional testing techniques are insufficient to test Big Data systems, but contemporary testing techniques, some of which are domain-specific and a combination of more than one techniques, are more efficient for testing Big Data. We study diverse functional, non-functional, and combined (functional and non-functional) testing techniques that have been used to solve the challenges faced during Big Data testing. We have also observed that these diverse techniques outperform traditional testing techniques to test Big Data, i.e., multi-agent-based testing, fault tolerance testing, model-driven entity reconciliation testing, fault identification (MRTree), MongoDB and XML based function testing, etc. We identified various challenges during the Big Data processing phases and highlighted the proposed solutions. Further, we also highlighted that the immense volume of data and a discrepancy in transformed and transmitted data are significant problems in Big Data processing testing. Future work can expand this work to identify test oracle problems, define generic test models and automated testing frameworks for Big Data testing. Other avenues of future work include Big Data testing for upcoming domains such as industry 4.0 and 5.0 systems and for data originating out of pandemics such as Covid-19.

**Funding Statement:** This research was supported by a research grant from Science Foundation Ireland (SFI) under Grant Number SFI/16/RC/3918 (Confirm) and Marie Sklodowska Curie grant agreement No. 847577 co-funded by the European Regional Development Fund. Wasif Afzal has received funding from the European Union's Horizon 2020 research and innovation program under grant agreement Nos. 871319, 957212; and from the ECSEL Joint Undertaking (JU) under grant agreement No 101007350.

**Conflicts of Interest:** The authors declare that they have no conflicts of interest to report regarding the present study.